\documentclass[11pt,a4paper]{article}

\usepackage[T1]{fontenc}
\usepackage[utf8]{inputenc}
\usepackage[english]{babel}
\usepackage{lmodern}
\usepackage[margin=1in]{geometry}

\usepackage{amsmath}
\usepackage{amssymb}
\usepackage{amsthm}
\usepackage{graphicx}
\usepackage{multirow}
\usepackage{caption}
\usepackage{subcaption}
\usepackage{authblk}
\usepackage{placeins}
\usepackage[numbers,sort&compress]{natbib}
\usepackage[colorlinks=true,allcolors=blue]{hyperref}

\graphicspath{{imgs/}}

\title{\bfseries The Role of Network Topology and Opponent Information in
Shaping Cooperation in Multi-Agent Reinforcement Learning Systems}

\author[1,2]{Seongho Son\thanks{Corresponding author: \texttt{seong.son.22@ucl.ac.uk}}}
\author[1]{Stephen Hailes}
\author[1,2,3]{Mirco Musolesi}
\affil[1]{Department of Computer Science, University College London}
\affil[2]{UCL Centre for Artificial Intelligence, University College London}
\affil[3]{Department of Computer Science and Engineering, University of Bologna}

\date{}

\begin{document}

\maketitle

\begin{abstract}
\noindent
Several works have investigated the influence of graph topology on cooperation among artificial agents, while the majority of the literature has focused on modelling agents' adaptation through strategy imitation, which relies solely on the cumulative payoffs of others. This paper investigates scenarios in which each agent learns to play the two-player Iterated Prisoner's Dilemma (IPD) using deep reinforcement learning. Each agent is represented as a node in a graph, where its neighbours constitute the pool of opponents with whom it can interact. During each IPD episode, agents are provided with different types of information about their opponent, consisting of action history and opponent identity. Experimental results across different graph topologies show that the number of neighbours per node and the average path length are the main factors affecting the emergence of cooperation. We also show that, while partner selection fosters mutual cooperation by limiting the diversity of the opponent pool, providing agents with the identity of their opponent hinders the proliferation of cooperative strategies.
\end{abstract}


\vspace{1em}

\section{Introduction}
\label{sec:intro}

A large portion of interactions within human societies consists of altruistic behaviour, ranging from donating blood \cite{ferguson2015mechanism,attari2014reasons} to investing in early-stage ventures with little expectation of direct returns \cite{klyver2017altruistic}. Similar forms of altruistic behaviour can also emerge among artificial agents, where agents may incur individual costs to benefit others or contribute to collective outcomes.

If self-interested behaviour of each individual is assumed, what could be the factors that affect their tendency to be cooperative? The emergence of cooperation among multiple agents has attracted a substantial amount of attention in academic research \cite{axelrod1981evolution,axelrod1984evolution}. Social dilemmas in particular have been extensively studied, as they present mixed-motive situations where individually optimal choices are to act selfishly while socially optimal choices are to be cooperative \cite{nowak1992tit,masuda2003spatial,ale2013evolution}.

Reinforcement learning (RL) has been widely used to model the learning dynamics of societies of artificial and human agents \cite{lowe2017multi,anastassacos2020partner,merhej2022cooperation,wong2023deep,DagMus25:Investigating,leung2026:learningtocooperate}. In RL, agents learn by changing behaviour to maximise their return, which is defined as the discounted sum of rewards. If the agent receives a reward $r_k \in \mathbb{R}$ at time step $k$ and the discount rate is denoted as $\gamma \in (0, 1]$, the return $G$ is defined by $G = \sum_{k=0}^\infty \gamma^k r_k$. With RL, we can investigate the behaviour of individual agents learning with various reward structures and different levels of access to information about the surrounding environment. The stationarity of the environment surrounding the agent is essential for the theoretical guarantee of the agent's performance improvement, which is violated with multiple agents learning at the same time \cite{wong2023deep}. Along with partial observability of the state, these features of multi-agent environments have been the major obstacles to stable and efficient training. Although approaches such as \cite{lowe2017multi,foerster2018learningopponentlearningawareness} tried to address this issue by modelling or actually gaining access to the information of the agents' opponents, the methods can become intractable due to weak scalability. On the other hand, partner selection algorithms have been proposed as an approach to training multiple RL agents in decentralised scenarios, enabling agents to learn norm-inducing behaviours and promote cooperation \cite{anastassacos2020partner,Koohborfardhaghighi2021effectofpartnerselection,Gu2023aflexipartnerselection,DagMus25:Investigating}.

Graph topology representing agents as nodes provides a useful way to analyse relation between individuals and draw insights from the structure. Numerous approaches have been presented on the issue of cooperation using graph representation of the society \cite{Santos2005ScalefreeNP,Santos2006evolutionary,santos2008social,rezaei2012dynamic, barlow13:impact,Hol2013spatialstructure,ishibuchi2013evolution,Cimini_2014,LIU2017156,bara2022enabling, song2026network,yuan2026dynamics}. Many existing works simulated agents' behaviour with learning algorithm based on imitation, in which each agent copies one of its neighbour's strategies with probability proportional to the difference between their accumulated payoffs. This approach enables agents to adopt the strategy that results in higher payoff, while it requires access to information on others' payoffs \cite{santos2008social,rezaei2012dynamic,LIU2017156,Santos2005ScalefreeNP,Santos2006evolutionary}. Reinforcement learning does not rely on this requirement by using only the reward gained by the agent itself to adjust its behaviour. Although previous studies have highlighted the significant impact of graph topology on the emergence of cooperation, relatively little research has explored how it shapes the behaviour of reinforcement learning--based agents.

During interactions, agents' behaviours strongly depend on the information about their opponents. Previous work has investigated the role of reputation in the emergence of cooperation, often in the format of action history \cite{anastassacos2020partner,axelrod1980effective,fu2008reputation,anastassacos2021cooperationreputationdynamicsreinforcement}. Agents can prefer other agents with cooperative action history as their interaction partners, which can motivate agents to build a reputation of being cooperative. On the other hand, exposing information on their behaviours can render agents vulnerable to be targeted by defective agents and be exploited. Agents' behaviours can also be affected by the presence of their interaction partners' identity information in their observation \cite{Brewer1986ChoiceBI,brent2006socialidentity,Aksoy2019CrosscuttingCI}. When agents can observe the identity of their interaction partners, they can use different strategy. Behaviours like unconditionally cooperating with some of the agents while exploiting the others are possible in this setting. In this paper, we investigate cases where identity information is provided such that each individual can be distinguished from others by decoding state information.

Our work investigates how graph topology and opponent identity influence the emergence of cooperation. Each agent is represented as a node in a graph and interacts with its neighbours by playing two-player Iterated Prisoner's Dilemma (IPD) \cite{axelrod1981evolution}. We provide experimental results with three different types of synthetic graphs and show that longer average path length and sparser graphs lead to more mutual cooperation of the agents. We also conduct experiments in which agents perform partner selection before IPD. We demonstrate that access to longer histories of an opponent's actions allows agents to more accurately discern the opponent's strategic behaviour. We report the results of the experiment of including identity information from the opponent in the state, in which the emergence of cooperation is hindered by enabling agents with defective strategies to spot entirely cooperative agents while avoiding retaliating agents, while providing more information on the action history alleviates the phenomenon.

\section{Iterated Prisoner's Dilemma on Graphs}
\label{sec:ipd}

\subsection{Game Structure}
\label{sec:game}

We assume that each agent in the population is represented as a node $v \in \mathcal{V}$ in an undirected graph $\mathcal{G} = (\mathcal{V}, \mathcal{E})$, with the set of edges in the graph denoted as $\mathcal{E}$. Each agent $i \in \{1, 2, ..., N\}$ participates in an interaction with one of the other agents whose nodes are in its neighbourhood $\mathcal{N}_v = \{u \,|\, (u, v) \in \mathcal{E} \vee (v, u) \in \mathcal{E}\}$. For each interaction agents decide whether to cooperate (C) or to defect (D), and receive payoffs $r$ based on the result of each interaction characterised by a payoff matrix of Prisoner's Dilemma (PD) \cite{axelrod1980effective,axelrod1981evolution,nowak1992tit,anastassacos2020partner}, as in Table~\ref{table:1}. A game is considered PD when $T > R > P > S$ and $2R > T + S$ are satisfied. We will denote 4 possible interactions in an episode of PD from the row player's perspective: (C, C) as mutual cooperation, (C, D) as sucker, (D, C) as temptation, (D, D) as mutual defection, where $(a_i, a_j)$ represents action pair of row player $i$ and column player $j$.

We further specify the relationship between payoffs using benefit to cost ratio $b:c = 5:1$ \cite{ale2013evolution}, where $R = b - c, T = b, S = -c, P = 0$. Payoff values are shifted such that $R + P = T + S = 0$, which makes initialised strategies of agents be equally distributed across cooperative and defective strategies. Payoff values are rescaled to prevent numerical instability during the training of neural networks.

\begin{table}[h!]
    \centering
    \begin{tabular}{|c|c|c|}
         \hline
         PD & C & D \\
         \hline
         C & (R, R) & (S, T) \\
         \hline
         D & (T, S) & (P, P) \\
         \hline
    \end{tabular}
\quad
    \begin{tabular}{|c|c|c|}
         \hline
         IPD & C & D \\
         \hline
         C & (0.2, 0.2) & (-0.3, 0.3) \\
         \hline
         D & (0.3, -0.3) & (-0.2, -0.2) \\
         \hline
    \end{tabular}
\caption{(a) General format of payoff matrix used for an episode of Prisoner's Dilemma (left). (b) Payoff matrix used in each interaction of Iterated Prisoner's Dilemma (right).}
\label{table:1}
\end{table}

Each agent $i$ is randomly assigned an opponent $o \in \mathcal{N}_i$ for an interaction. Due to the randomness of the assignment of the opponent and the structure of the graph, each agent can interact with different numbers of opponents in an interaction round $k$. We index the interactions occurring in each round as $t \in \{1, ..., N-1\}$. Agents are provided with information of the previous actions of the assigned opponents. The action history of the opponent is given in the format of a concatenated one hot vector, $s^i \in \mathcal{S}_{dil} = \mathbb{R}^{2 \times l}$, where $l$ denotes the length of the action history. This is used for the state vector of the agent's dilemma-playing Deep Q-network $f_{dil}^i(s^i)$ \cite{mnih2015human} to select an action $a^i \in \mathcal{A}_{dil} = \{C, D\}$. We use $l = 1$, and report the result of the experiment after running each setting for 100,000 interaction rounds.

\subsection{Synthetic Graphs}
\label{sec:graphs}

For experiments with random assignment of opponents, we use three graph types: Erd\H{o}s-R\'enyi (ER) \cite{erd6s1960evolution}, Watts-Strogatz (WS) \cite{watts1998collective} and Barab\'asi-Albert (BA) \cite{barabasi1999emergence}. For ER graphs, we generate graphs with wiring probability $p_{ER}$ sampled from $U[0.1, 1]$. For WS graphs, we first construct a regular graph with 4 edges per node and rewire edges with probability $p_{WS} \sim U[0, 1]$. For BA graphs, we denote the number of edges added per each preferential attachment as $m \in \{1, 2, 3, 4\}$. In BA graphs, preferential attachment is performed after initial $m$ nodes.

We generate 100 graphs for each type with parameters sampled with uniform probability using the method mentioned above. For experiments with partner selection, we only use graphs with full connectivity. For all experiments, we use $N = 32$ agents for the population.

\subsection{IPD with Partner Selection}
\label{sec:ipd-ps}

Inspired by \cite{anastassacos2020partner}, we conduct experiments on cases where partner selection is involved. Each agent $i \in \{1, ..., N\}$ gets an opportunity to select a partner using its partner selection module $\eta_{sel, \pi}^i$. The input for this module consists of the concatenated state vector of neighbouring $(N-1)$ agents in the graph, $S_{sel}^i \in \mathbb{R}^{(N-1) \times d_{dil}}$. $d_{dil}$ denotes the dimension of each state vector used for each interaction of Prisoner's Dilemma. The output of the partner selection module consists of selection probability for each candidate, $\eta_{sel, \pi}^i(S_{sel}^i) \in \mathbb{R}^{(N-1)}$.

For the state vectors $S_{sel}^i$ used for playing IPD in experiments with partner selection, we add an option to provide the identity of the opponent. The opponent's index $o \in \{1, ..., i-1, i+1, ..., N\}$ is provided in the format of binary encoding. For example, in the population with 32 agents, an opponent with index number 13 will be represented as $01101_{2}$. This makes $d_{dil} = 2 \times l + \lceil \log_2 N \rceil$ when opponent identity is included. We run experiments with 6 different settings, where only one of two settings with the same value $l \in \{1, 5, 10\}$ has opponent identity in the dilemma-playing state $s$. Each setting is run with 20 different seeds, with each run consisting of 200,000 interaction rounds.

\section{Reinforcement Learning Implementation}
\label{sec:rl}

\subsection{Playing the IPD}
\label{sec:rl-ipd}

The learning algorithm used for this experiment is a variant of Deep Q-learning \cite{mnih2015human}. Deep Q-learning uses neural networks to estimate the optimal action-value function $Q^*$. When the agent selects an action $a_t$ using policy $\pi$ and receives reward $r_t$ for the action, and when the discount rate is $\gamma$, $Q^*(s_t, a_t) = \max_{\pi}\mathbb{E}[\sum_{j = 0}^\infty \gamma^{j} r_{t + j} | \pi]$. Each agent is equipped with a separate neural network for Prisoner's Dilemma $f^i_{dil}$ with its parameter denoted as $\theta_{dil}^i$. Agents are trained independently with each of them having a buffer to store transitions $\{(s_{k, t}, a_{k, t}, r_{k, t}, s_{k, t+1}): k = 1, ..., K, t = 1, ..., N-1\}$ used for training. The policy of each agent is updated after each interaction round $k$ and transitions are discarded from training buffer after each update, in order to maintain relevance to the current transition dynamics of the environment \cite{anastassacos2020partner}. The dilemma-playing policy of agent $i$ can be described as
\begin{align}\label{eq:policy}
\pi^i(s^i) =
\begin{cases}
    \text{argmax}_{a^i \in \mathcal{A}_{dil}} Q^i(s^i, a^i)& \text{with probability} \ 1 - \epsilon\\
    \mathcal{U}(\mathcal{A}_{dil}),              & \text{otherwise,}
\end{cases}
\end{align}

\noindent where $\mathcal{U}(\mathcal{A}_{dil})$ denotes sampling from a uniform distribution \cite{anastassacos2020partner}. After each interaction round $k$, dilemma-playing networks update their Q-values by minimising the objective function $L_k(\theta_{dil, k}^i)$ as described below. When calculating the objective function, we estimate the Q-value in the next time step $t+1$ with another network called the target network to stabilise learning \cite{mnih2015human}. The target network is initialised with the same architecture and parameter values as the dilemma-playing network, while its parameters $\theta^{i, -}_{dil}$ are copied from the original network every $Z = 16$ interaction rounds. We use exploration rate $\epsilon = 0.05$, discount rate $\gamma = 0.99$. We use deep neural networks with two hidden layers, 32 and 16 nodes respectively. Rectified Linear Unit is used for the activation function. With $s'$ denoting the state at the next time step $s_{k, t+1}$, the loss function is defined as follows:

\begin{align}\label{eq:dqn-loss}
\begin{split}
L_k(\theta_{dil, k}^i) = \mathbb{E}_{s, a, r, s'}\biggl[\bigl(y^i_{k, t} - Q^i(s^i_{k, t}, a^i_{k, t}; \theta_{dil, k}^i)\bigr)^2\biggr],\\
y^i_{k, t} = r^i_{k, t} + \gamma \max_{a' \in \mathcal{A}_{dil}} Q^i(s^i_{k, t+1}, a'; \theta^{i, -}_{dil, k})
\end{split}
\end{align}

\begin{figure}[!htbp]
    \centering
    \includegraphics[
        width=0.75\textwidth,
        trim={2cm 4cm 4cm 2cm}, clip
    ]{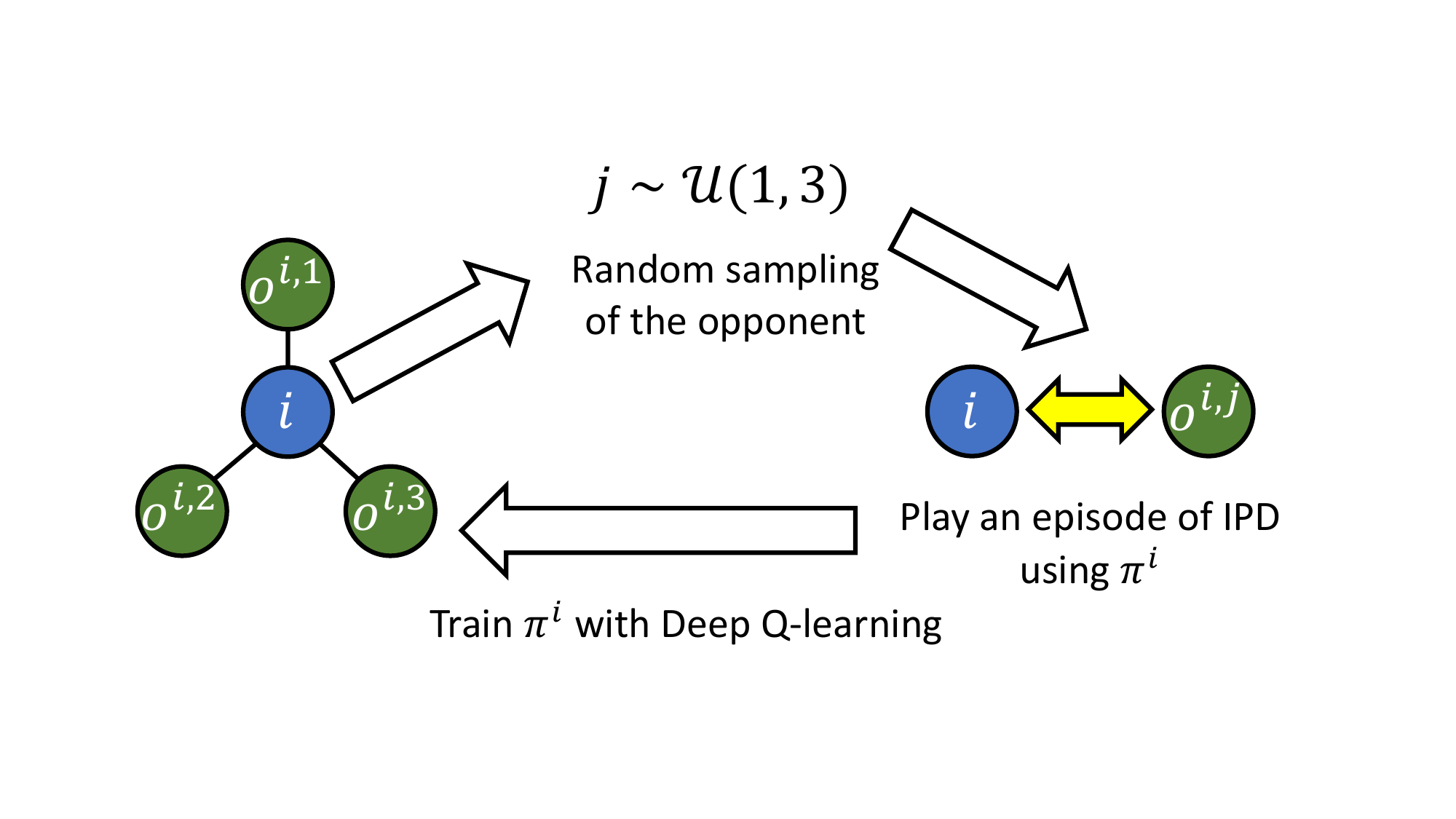}
    \caption{Visual summary of IPD on graphs, from the perspective of $i$th agent. For each interaction round, each agent selects an opponent among its neighbours in the graph with uniform probability. Both experiences of playing with opponents that agent $i$ selected or the agent being selected by one of its neighbours are used for training the dilemma-playing policy $\pi^i$.}
    \label{fig:scheme-ipd}
\end{figure}

Figure~\ref{fig:scheme-ipd} summarises the resulting interaction and training loop from the perspective of agent $i$.

\subsection{Partner Selection}
\label{sec:rl-ps}

\begin{figure}[!htbp]
    \centering
    \includegraphics[
        width=0.75\textwidth,
        trim={3cm 1.5cm 3cm 0.5cm}, clip
    ]{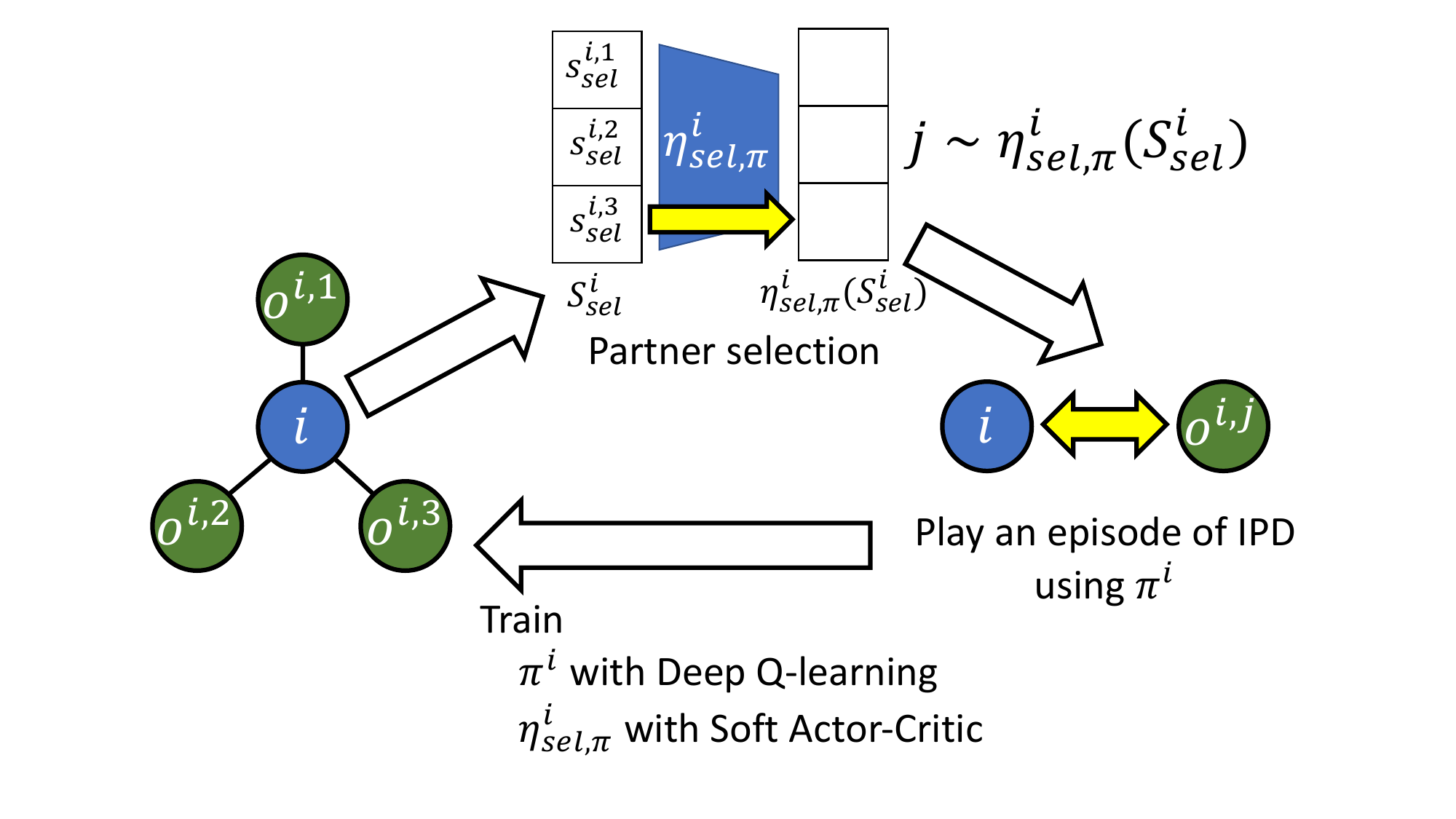}
    \caption{Visual summary of IPD with partner selection, from the perspective of $i$th agent. For each interaction round, each agent selects an opponent among the other $N-1$ agents using partner selection module $\eta^i_{sel, \pi}$. The partner selection module is trained with only the experience of agent $i$ selecting an opponent, as the action of partner selection $a^i_{sel}$ is the index of the selected opponent.}
    \label{fig:scheme-ps}
\end{figure}

In experiments with partner selection, each agent uses separate neural networks for learning how to choose the opponent. These networks used for selection $f_{sel, Q}^i, f_{sel, \pi}^i$ have the same structure as the dilemma-playing network, except that their outputs are single values representing the Q-value or action probability in $\mathbb{R}^1$ for selecting the given agent as a partner. With the information of agent $i$'s neighbour $o^{i, j} \in \mathcal{N}_i, j \in \{1, ..., |\mathcal{N}_i|\}$ denoted as $s_{sel}^{i, j}$, $S_{sel}^i$ is $[s_{sel}^{i, 1}, s_{sel}^{i, 2}, ..., s_{sel}^{i, |\mathcal{N}_i|}]$, and selection network is applied to each of $s_{sel}^{i, j}$. The outputs of the partner selection module then become $\eta_{sel, Q}^{i}(S_{sel}^i) = [f_{sel, Q}^i(s_{sel}^{i, 1}), f_{sel, Q}^i(s_{sel}^{i, 2}), ..., f_{sel, Q}^i(s_{sel}^{i, |\mathcal{N}_i|})]$ for the Q-function and $\eta_{sel, \pi}^{i}(S_{sel}^i) = [f_{sel, \pi}^i(s_{sel}^{i, 1}), f_{sel, \pi}^i(s_{sel}^{i, 2}), ..., f_{sel, \pi}^i(s_{sel}^{i, |\mathcal{N}_i|})]$ for the policy. We use the information from agent $i$'s neighbours at $t = 1$ for $s_{sel}^{i, j}$ in the interaction round $k$, because each agent performs partner selection before any agent in the population plays an episode of the Prisoner's Dilemma. Figure~\ref{fig:scheme-ps} summarises the corresponding procedure when partner selection is involved.

We use Soft Actor-Critic (SAC) \cite{haarnoja2019softactorcriticalgorithmsapplications} for training the partner selection networks. SAC adds an entropy term $\mathbb{H}(\pi(\cdot|s))$ as a regularisation term to the policy gradient objective, making the optimal policy $\pi^* = \text{argmax}_\pi \sum_{k=0}^K \mathbb{E}_{s_t \sim p, a_t \sim \pi}[r(s_t, a_t) + \alpha \mathbb{H}(\pi(\cdot|s_t))]$ and $\mathbb{H}(\pi(\cdot|s)) = - \int \pi(a|s) \log\pi(a|s)da = \mathbb{E}_{a \sim \pi(\cdot|s)}[-\log\pi(a|s)]$. $\alpha = 0.01$ is the parameter determining the importance of the entropy term against the reward. The soft state-value function and soft Q-function are defined as $V(s) = \mathbb{E}_{a \sim \pi}[Q(s, a) - \alpha \log(\pi(a|s))]$. We denote the parameters of agent $i$'s selection policy as $\phi^i_{sel}$, and parameters of the soft Q-function for selection as $\theta_{sel}^i$. We also use target network parameters $\bar{\theta}_{sel}$ for soft Q-function like we did for training policies to play IPD. $r^i_{k}$ denotes the payoff gained by playing with the opponent which agent $i$ selected in round $k$. $\theta_{sel}$ are trained to minimise the soft Bellman residual:

\begin{align}\label{eq:sac-q}
\begin{split}
L_{sel, k}^{Q}(\theta_{sel, k}^i) = \mathbb{E}_{s, a, r, s'}\biggl[\bigl(Q_{sel}^i(s, a; \theta_{sel, k}^i) - y^i_{sel, k}\bigr)^2\biggr],\\
y^i_{sel, k} = r^i_{sel, k} + \gamma\sum_{a'}\pi(a'|s'; \phi_{sel, k}^i)(Q_{sel}^i(s', a'; \bar{\theta}_{sel, k}^i) - \alpha \log \pi(a'|s'; \phi_{sel, k}^i)), \\
s = s^i_{sel, k},\ a = a^i_{sel, k},\ r = r^i_{sel, k},\ s' = s^i_{sel, k+1},\ a' = a^i_{sel, k+1}.
\end{split}
\end{align}

In SAC, policies are updated towards the exponential of the soft Q-function:

\begin{align}\label{eq:sac-pi}
\begin{split}
L_{sel, k}^{\pi}(\phi_{sel, k}^i) = \mathbb{E}_{s}\biggl[
\sum_a\pi(a|s; \phi_{sel, k}^i)(
\alpha \log \pi(a|s; \phi_{sel, k}^i) -
Q_{sel}^i(s, a; \theta_{sel, k}^i)
)
\biggr],\\
s = s^i_{sel, k},\ a = a^i_{sel, k}.
\end{split}
\end{align}

For target networks of soft Q-functions, we use $\tau = 0.05$ to gradually apply update: $\bar{\theta}_{sel, k}^i \leftarrow \tau \theta_{sel, k}^i + (1 - \tau) \bar{\theta}_{sel, k}^i$. This implementation of partner selection is different from that of \cite{anastassacos2020partner}, which gave implicit information on opponent identity to only selection networks. Our architecture enables us to have the option of explicitly including identity information in the state vectors and investigate the effect of it on the interaction dynamics of playing IPD. The implementation of the environment and of both learning algorithms, together with the configuration files reproducing every experiment reported below, is publicly available.

\section{Results}
\label{sec:results}

\subsection{IPD with Random Matching}
\label{sec:res-random}

\begin{figure}[!htbp]
    \centering
    \includegraphics[width=0.32\textwidth]{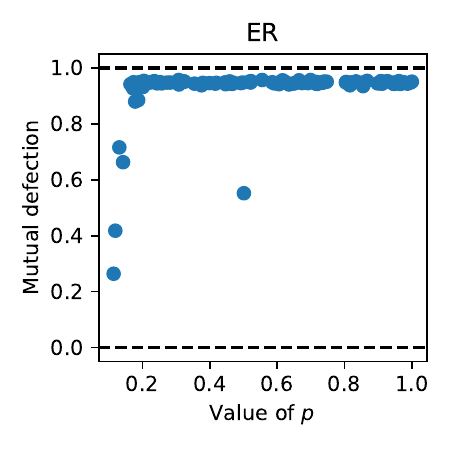}
    \includegraphics[width=0.32\textwidth]{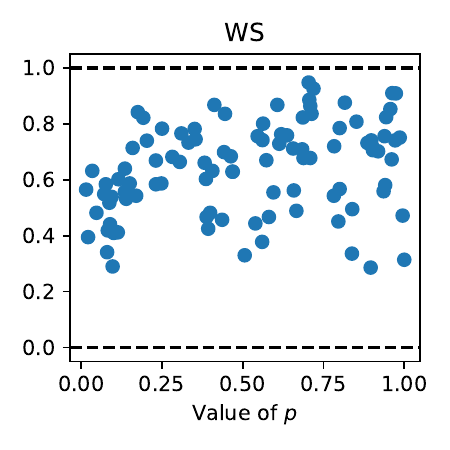}
    \includegraphics[width=0.32\textwidth]{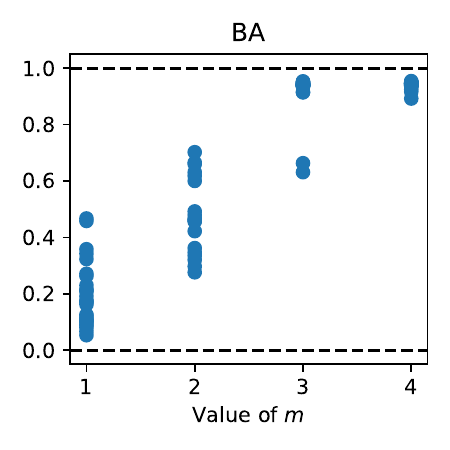}
    \caption{Experimental results per graph type with parameters for graph generation (ER, WS and BA from the left, respectively).}
    \label{fig:netparam-dd}
\end{figure}

We first present in Figure~\ref{fig:netparam-dd} the proportion of mutual defection $p_{md}$ in each run's last 100 interaction rounds with the parameters used to generate graphs. Most of the runs on ER graphs with $p_{ER} > 0.2$ converge to complete mutual defection. In WS graphs the tendency is less apparent compared to that of ER graphs, but as the value of $p_{WS}$ increases, the higher the level of mutual defection. In BA graphs, as the value of $m$ increases, runs tend to converge to a higher level of mutual defection.

\begin{figure}[!htbp]
    \centering
    \includegraphics[width=0.32\textwidth]{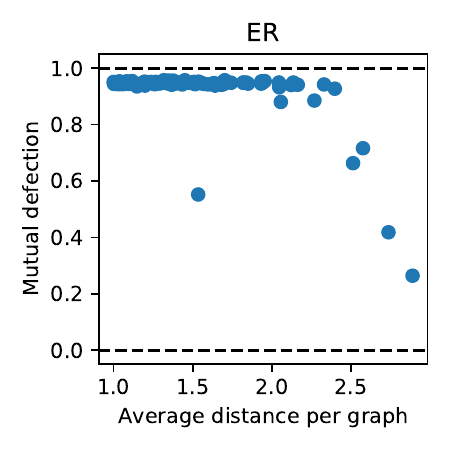}
    \includegraphics[width=0.32\textwidth]{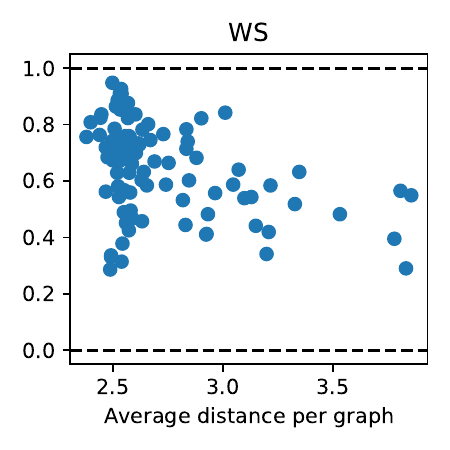}
    \includegraphics[width=0.32\textwidth]{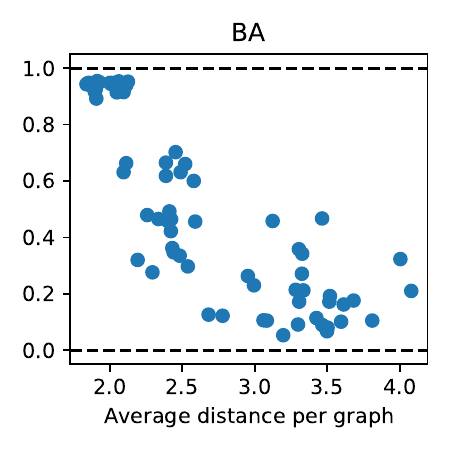}
    \caption{Experimental results per graph type with average path length per graph (ER, WS and BA from the left, respectively).}
    \label{fig:avgdist-dd}
\end{figure}

\begin{figure}[!htbp]
    \centering
    \includegraphics[width=0.32\textwidth]{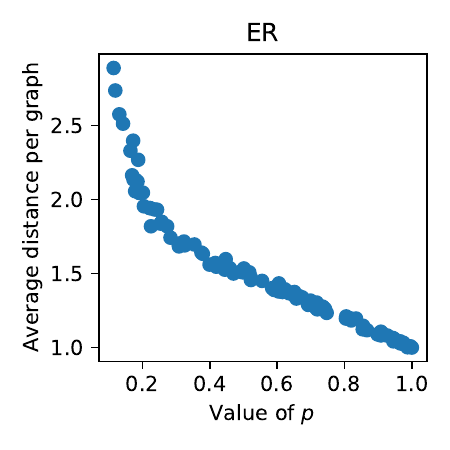}
    \includegraphics[width=0.32\textwidth]{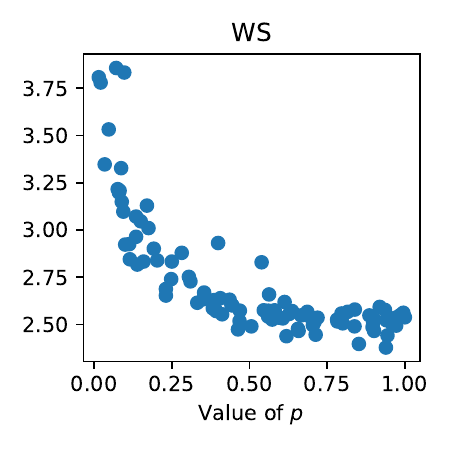}
    \includegraphics[width=0.32\textwidth]{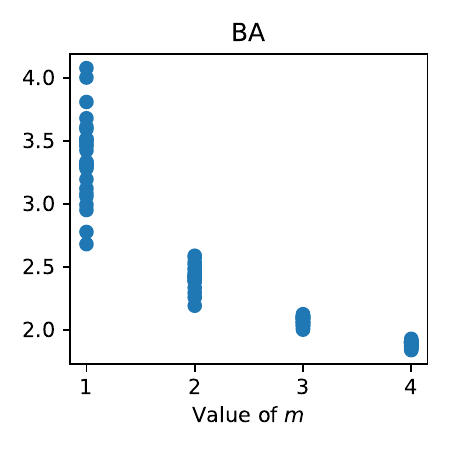}
    \caption{Relation between graph generation parameters and average path length (ER, WS and BA from the left, respectively).}
    \label{fig:netparam-avgdist}
\end{figure}

We further investigate this phenomenon using metrics calculated from graphs. Figure~\ref{fig:avgdist-dd} shows the results of the experiment considering the relationship with the average path length, which is calculated by averaging the shortest path length between each pair of nodes in the graph: $d_{path}(\mathcal{G}) = \frac{\sum_{v, u \in \mathcal{V}, v \neq u}d(v, u)}{|\mathcal{V}|(|\mathcal{V}|-1)}$ where $d(v, u)$ denotes the shortest path length between vertices $v$ and $u$. Figure~\ref{fig:netparam-avgdist} shows the relation between the graph generation parameters $p_{ER}, p_{WS}, m$ and the average path length of each generated graph. Across all three types of graphs, more runs converge to a higher value of $p_{md}$ as the average path length decreases. In ER graphs, except one outlier, the other runs with $d_{path} < 2.5$ converge to $p_{md} > 0.8$. BA graphs are characterised by a longer $d_{path}$ than ER graphs, reaching $d_{path} > 4$ when $m = 1$. The level of mutual defection in BA graphs with $d_{path} > 3$ does not increase above 0.6, while many graphs generated with $d_{path} < 2.5$ result in $p_{md} > 0.8$. This corresponds to cases where $m = 3, 4$ as shown in Figure~\ref{fig:netparam-dd}. A similar pattern is shown in the results of the WS networks, with the value of $p_{md}$ staying below 0.6 when $d_{path} > 3.5$ and reaching above 0.9 when $d_{path} < 3$.

\subsection{IPD with Partner Selection}
\label{sec:res-ps}

\begin{figure}[!htbp]
    \centering
    \includegraphics[width=0.6\textwidth]{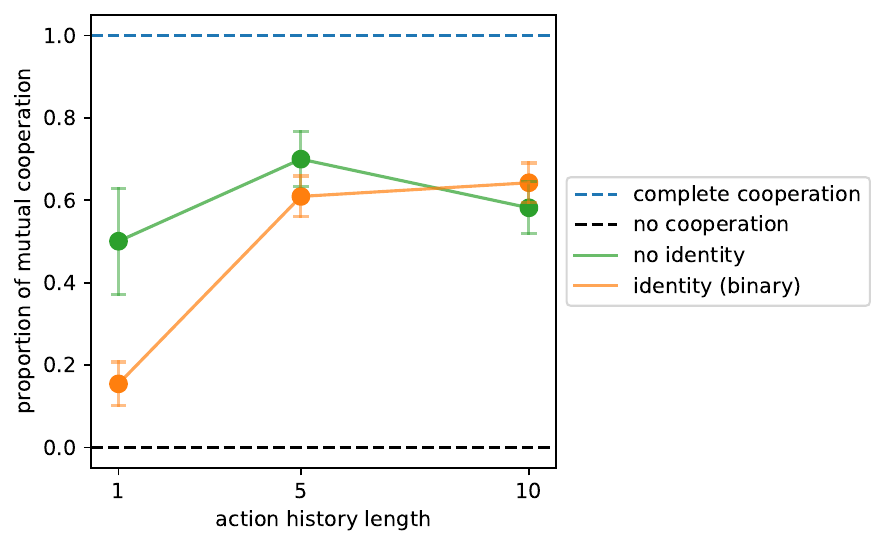}
    \caption{Experimental results of partner selection experiment. Dots represent the average proportion of mutual cooperation in the last 100 interaction rounds of 20 runs, while vertical intervals represent 95\% confidence intervals of each experiment setting. Experiments conducted without having identity information in the states of dilemma-playing agents are coloured in green, while experiments which included binary encoding of the opponent's identity index are coloured in orange. Except for the experiment with identity information and action history length $l = 1$, which showed significantly lower level of mutual cooperation, all the other experiments' confidence intervals overlap.}
    \label{fig:ps-compare}
\end{figure}

Figure~\ref{fig:ps-compare} shows the result of the experiments performed with partner selection. The average proportion of mutual cooperation in the last 100 interaction rounds, denoted as $p_{mc}$, is calculated across 20 runs with 95\% confidence interval. When no information on the identity of the opponent is included in the agent state information, the level of mutual cooperation after 200,000 interaction rounds does not show a statistically significant difference when the length of the action history $l$ changes. Significant difference occurs when binary encoding of the opponent's index is provided in the state information. $p_{mc}$ drops below 0.3 when $l = 1$, largely deviating from cases without the identity of the opponent. $p_{mc}$ increases above 0.5 when $l = 5, 10$, which makes the confidence intervals of $p_{mc}$ overlap with the other experiments without identity information. This implies that the effect of identity information on the proportion of mutual cooperation becomes insignificant as action history length increases.

\begin{figure}[!htbp]
    \centering
    \includegraphics[width=\textwidth]{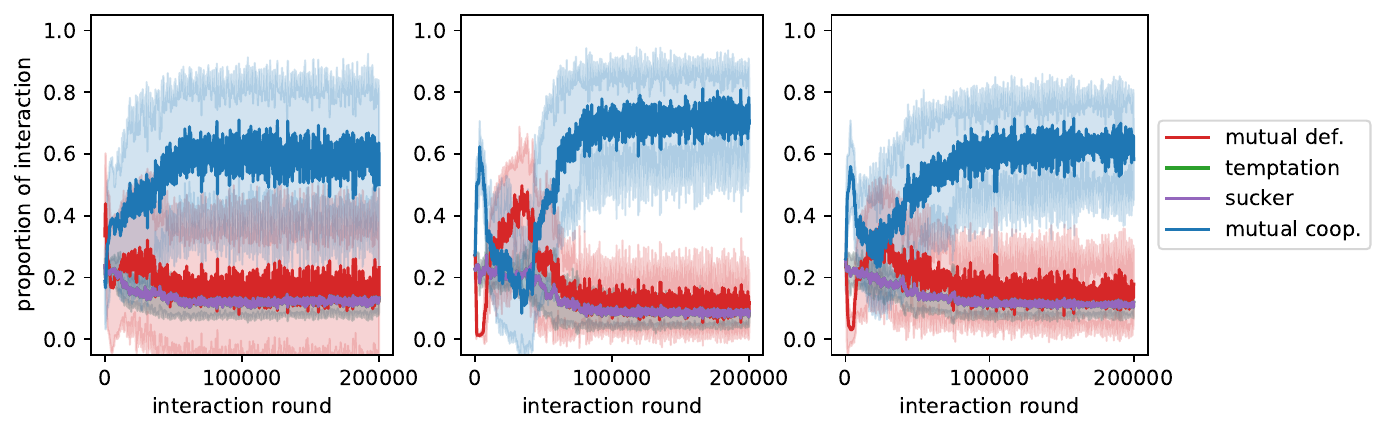}
    \caption{Learning dynamics of partner selection experiment without opponent identity ($l = 1, 5, 10$ from the left, respectively).}
    \label{fig:dyn-noid}
\end{figure}

\begin{figure}[!htbp]
    \centering
    \includegraphics[width=\textwidth]{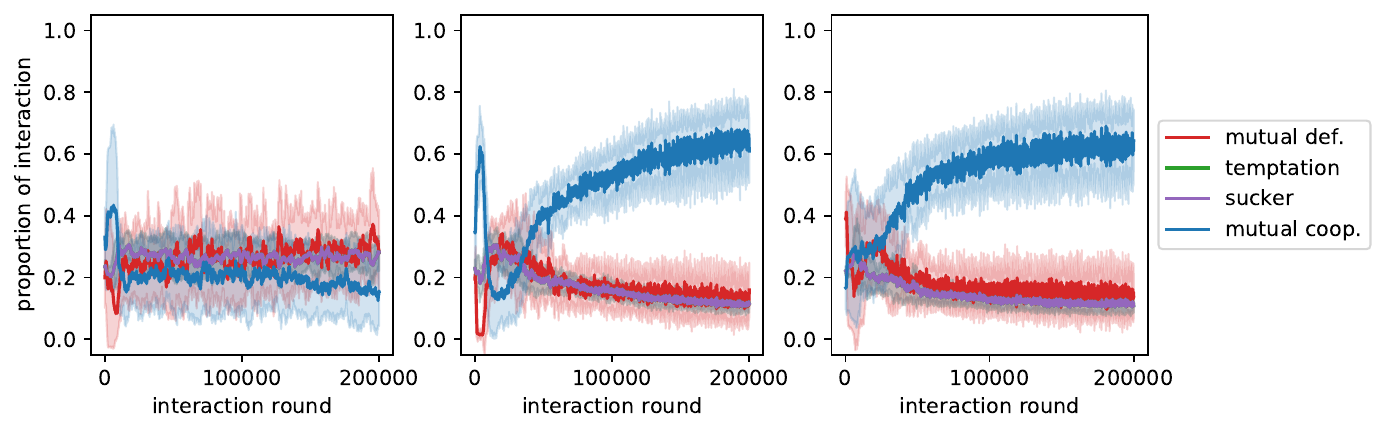}
    \caption{Learning dynamics of partner selection experiment with opponent identity ($l = 1, 5, 10$ from the left, respectively).}
    \label{fig:dyn-id}
\end{figure}

We also investigate the learning dynamics by measuring the proportion of four types of interactions between agents in Iterated Prisoner's Dilemma every 100 interaction rounds. Shaded areas show the standard deviation of interactions calculated across 20 runs. Figure~\ref{fig:dyn-noid} shows the learning dynamics of the partner selection experiments without the opponent identity provided. At $l = 1$, mutual cooperation spreads rapidly during the initial 50,000 rounds of interaction and maintains its level afterwards. For $l = 5$ and $l = 10$, we observe learning dynamics similar to those reported in \cite{anastassacos2020partner}: (1) mutual cooperation initially increases in the population; (2) unilateral exploitation of cooperative agents (i.e., sucker and temptation outcomes) increases, while mutual cooperation decreases; (3) mutual defection increases, while unilateral exploitation decreases; and (4) mutual cooperation increases again, accompanied by decreases in both unilateral exploitation and mutual defection. Mutual cooperation reaches its highest level more slowly than for $l = 1$, with the peak occurring at around 100,000 rounds across the experimental runs for both $l = 5$ and $l = 10$.

In Figure~\ref{fig:dyn-id} we present the learning dynamics of experiments with opponent identity in the state information. At $l = 1$, the level of mutual cooperation increases in the initial stage to 0.4, but then decreases to 0.2 and gradually continues to decrease. All the other interactions, such as mutual defection or temptation, gradually increase throughout the experiment. However, with $l = 5$ and $l = 10$, mutual cooperation shows a stable increase after the first quarter of the experiments. The learning dynamics described in \cite{anastassacos2020partner} is also apparent, except that in $l = 10$, the initial surge of mutual cooperation is less significant.

\FloatBarrier

\section{Discussion}
\label{sec:discussion}

\subsection{Effect of Graph Topology on the Emergence of Cooperation}
\label{sec:disc-topology}

We discuss here how differences in the graph structure influenced the strategy of RL-based agents to play IPD. For effective discussion, we consider cases where $l = 1$ and opponent identity is excluded from state information. Because $|\mathcal{A}_{dil}| = 2$ and $|\mathcal{S}_{dil}| = 2$, there are 4 possible strategies determined by the difference of Q-value of the dilemma-playing network as in Table~\ref{table:2}. $s_C, s_D$ denote the states where the assigned opponent's previous action was cooperation and defection, respectively.

\begin{table}[htbp]
    \centering
    \begin{tabular}{@{}|c|c|c|}
         \hline
          & $Q(s_C, C) > Q(s_C, D)$ & $Q(s_C, C) \leq Q(s_C, D)$ \\
         \hline
         $Q(s_D, C) > Q(s_D, D)$ & cooperator & reverser\\
         \hline
         $Q(s_D, C) \leq Q(s_D, D)$ & retaliator & defector\\
        \hline
    \end{tabular}
    \caption{Possible strategies in experiments where $l = 1$, opponent identity excluded from state information.}
    \label{table:2}
\end{table}

It is well established that, in a single-shot Prisoner's Dilemma, defection is the optimal strategy for rational agents \cite{sandholm1996multiagent}. Previous studies show that in the iterated Prisoner's Dilemma, the discount factor $\gamma$ plays a significant role in the emergence of cooperation \cite{axelrod1984evolution,kreps1990course,fudenberg1991game}. In two-player IPD with indefinite length of episodes, when the opponent's strategy is tit-for-tat, the optimal strategy differs based on the value of $\gamma$ \cite{axelrod1984evolution}:

\begin{align}\label{eq:returns}
\begin{split}
&\text{Always cooperate:}\qquad G_c = \frac{R}{1 - \gamma},\\
&\text{Alternate between cooperation and defection:}\qquad G_a = \frac{T + \gamma S}{1 - \gamma^2},\\
&\text{Always defect:}\qquad G_d = T + \frac{\gamma P}{1 - \gamma}.
\end{split}
\end{align}

Using the payoff values we defined in Section~\ref{sec:game} $(T = 0.3, R = 0.2, P = -0.2, S = -0.3)$ and when $\gamma = 0.99$, values of the returns in each case above become $G_c = 20, G_a = 0.151, G_d = -19.5$. This shows that it is optimal to always cooperate when facing an agent with tit-for-tat strategy. If we ignore the actions randomly selected by $\epsilon$-greedy exploration, the agents with retaliator strategy in our setting shows the behaviour identical to that of ones with tit-for-tat strategy. Hence, we can say that agents continuously playing against opponents with retaliator strategy will eventually shift to cooperator strategy. Meanwhile, regardless of the value of discount factor, it is always optimal to defect against opponents with fixed strategy of cooperator, defector, and reverser.

\begin{figure}[!htbp]
    \centering
    \includegraphics[width=0.9\textwidth, trim={8cm 0.2cm 8cm 0}]{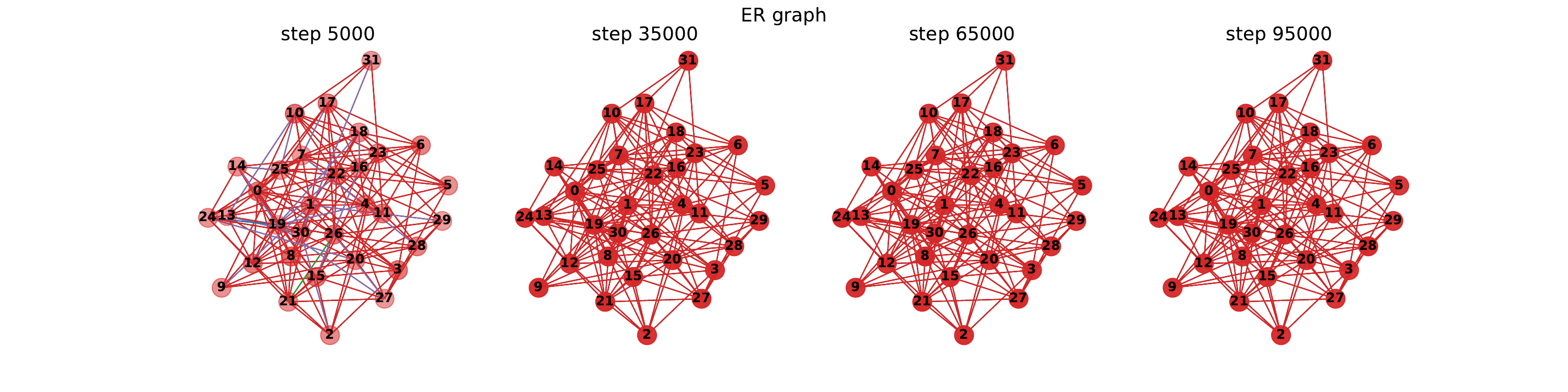}
    \includegraphics[width=0.9\textwidth, trim={8cm 0.2cm 8cm 0}]{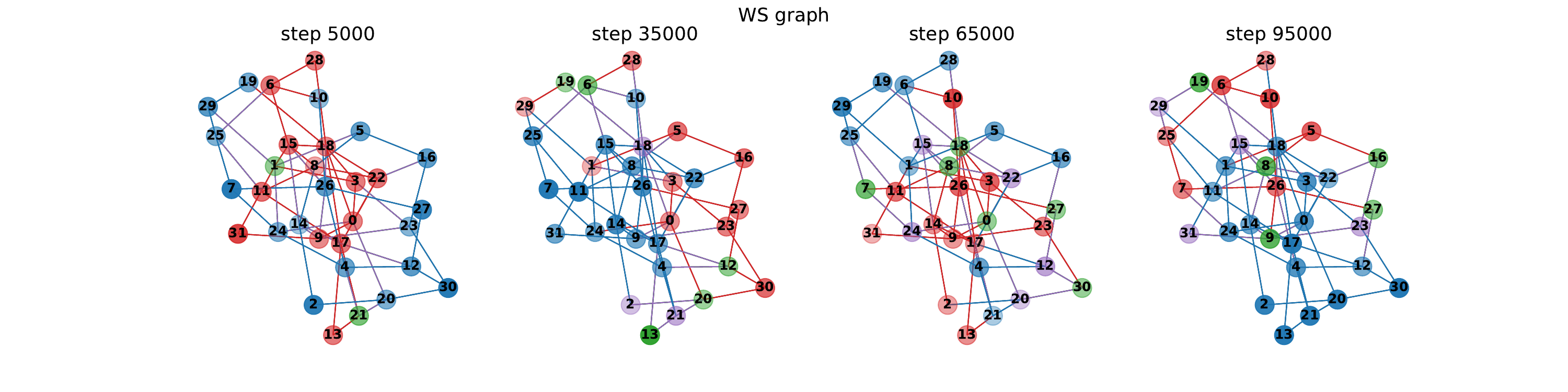}
    \includegraphics[width=0.9\textwidth, trim={8cm 0 8cm 0}]{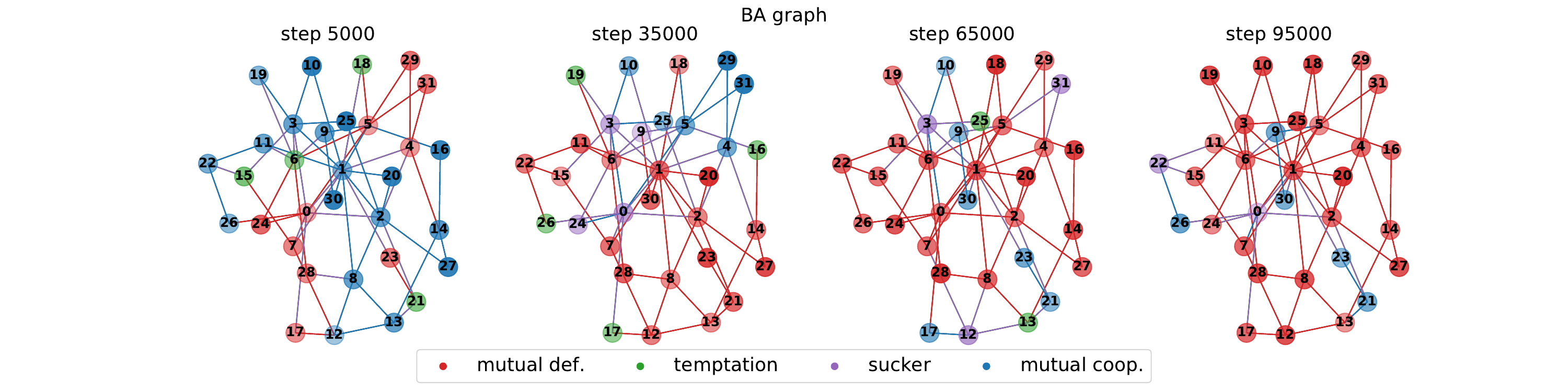}
    \caption{Interaction patterns in the IPD across different graph structures (ER, WS and BA from the top, respectively). Each graph shows results of one of the runs in each experiment setting. Each node represents an agent in the population. Colours of the nodes represent types of interactions the agent has experienced the most in the last 100 interaction rounds. Colours of the edges represent types of interactions that have occurred the most between the two nodes each edge is connecting.}
    \label{fig:netplot-graphs}
\end{figure}

In our actual experiments of random opponent assignment, agents are continuously learning from their recent experiences, which leads to change of their strategies. In addition, because the opponent for each interaction is assigned randomly from each agent's neighbours $\mathcal{N}$, agents are more likely to encounter an opponent with different strategy from that of the previous interaction when there are more neighbouring agents. This means that the expected number of interaction rounds of continuously playing with opponents with the same strategy gets smaller as $|\mathcal{N}|$ increases, which has similar effects to decreasing the discount factor $\gamma$. The gap of values between the returns $G_c, G_a, G_d$ defined above decreases as $\gamma$ decreases from 0.99, and three values become identical when $\gamma = 0.2$, which will make agents no longer continuously cooperate with tit-for-tat agents. Thus, we can say that an agent with larger neighbourhood in the graph is more likely to prefer defection. This prediction is verified from the results of ER and BA graphs, where increasing the value of graph generation parameter $p_{ER}$ and $m$ results in graphs with denser connection, also enlarging the number of edges in the graph and thus increasing the overall size of neighbourhood of all the agents in the graph. Figure~\ref{fig:netplot-graphs} visualises the resulting interaction patterns for one run of each graph type.

While the analysis based on neighbourhood size can explain the results of ER and BA graphs, it cannot be applied to that of WS graphs. In our experiments, we use WS graphs with fixed number of edges, only varying the rewiring probability $p_{WS}$. This means that the average size of the neighbourhood stays the same across all 100 runs. The factor that differs between each generated network is average path length, which decreases as $p_{WS}$ increases. With agents only interacting with others in adjacent nodes and changing its strategy through RL algorithms, the change of dilemma-playing strategy of an agent affects its neighbours first. With the same size of population and same value of average node degree, shorter average path length means that it needs less number of interaction rounds for a strategy to be propagated to all the other nodes. This way, defector strategy is able to propagate faster to the population in graphs with shorter average path lengths. Average path length also decreases as graph generation parameters $p_{ER}, m$ increase in ER and BA graph, which would further accelerate the proliferation of defective strategy in the population.

\subsection{Effect of Opponent Information on the Emergence of Cooperation}
\label{sec:disc-info}

We now discuss the effect of different formats of opponent information on the experiments with partner selection. In the experiments conducted by \cite{anastassacos2020partner}, agents were given action histories of opponents with $l = 1$, while selection module was implicitly aware of opponent identity. Partner selection played a crucial role in the emergence of cooperation, by enabling agents with cooperative strategies to choose other cooperators to benefit from mutual cooperation. Others with defective strategies also selected agents with cooperative history but were often retaliated against, which resulted in their shift to cooperative strategies as analysed in Section~\ref{sec:disc-topology}.

\begin{figure}[!htbp]
    \centering
    \includegraphics[width=0.9\textwidth, trim={8cm 0.2cm 8cm 0}]{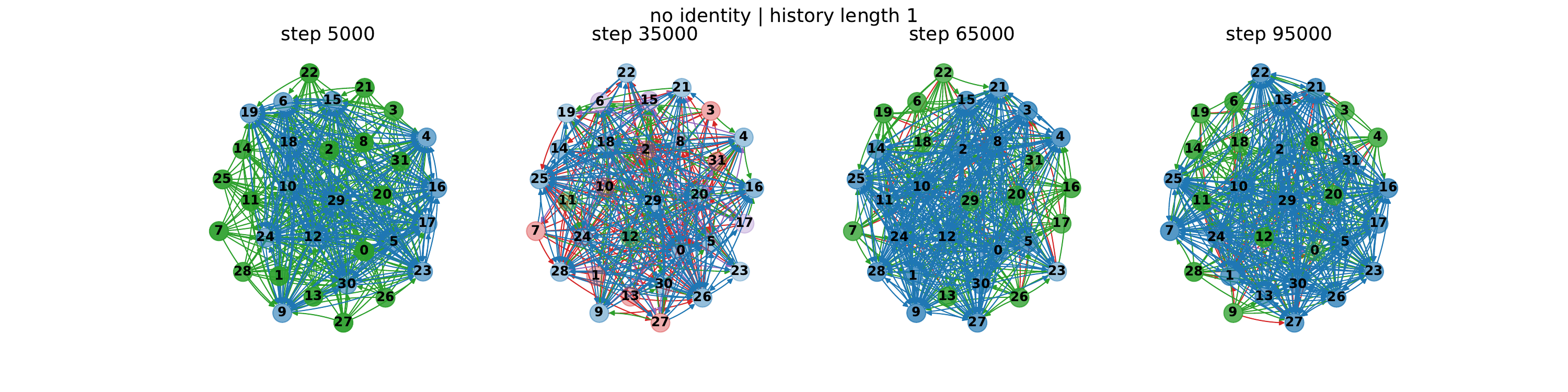}
    \includegraphics[width=0.9\textwidth, trim={8cm 0.2cm 8cm 0}]{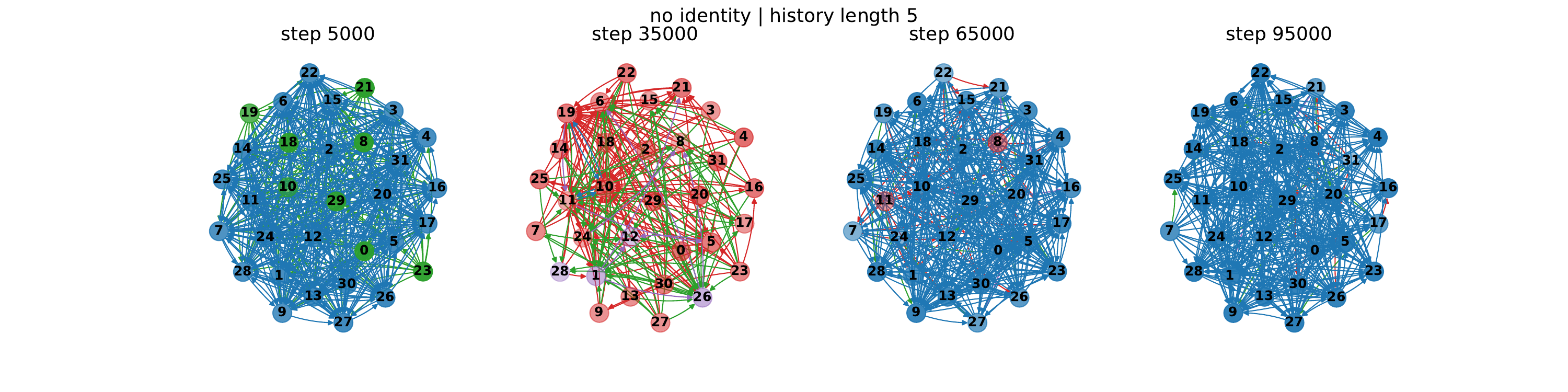}
    \includegraphics[width=0.9\textwidth, trim={8cm 0 8cm 0}]{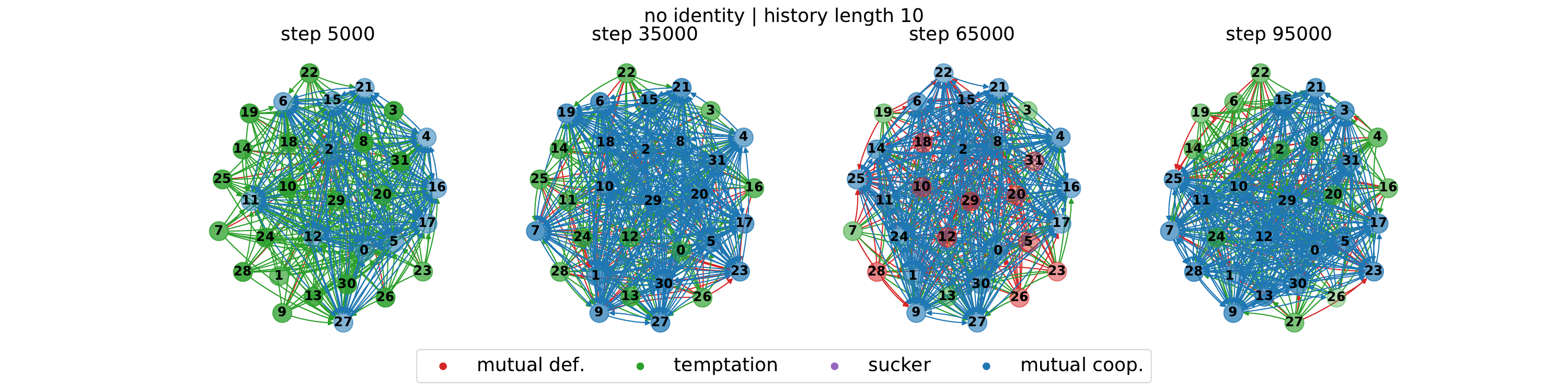}
    \caption{Interaction patterns in the IPD with partner selection without identity ($l = 1, 5, 10$ from the top, respectively). Each graph shows results of one of the 20 runs in each experiment setting. Edges have directions as each agent can select their interaction partners. Agents that mainly cooperate playing IPD (blue, purple nodes) are being selected from other agents, while agents that mainly exploit others (green nodes) are not being selected.}
    \label{fig:netplot-ps}
\end{figure}

\begin{figure}[!htbp]
    \centering
    \includegraphics[width=0.8\textwidth]{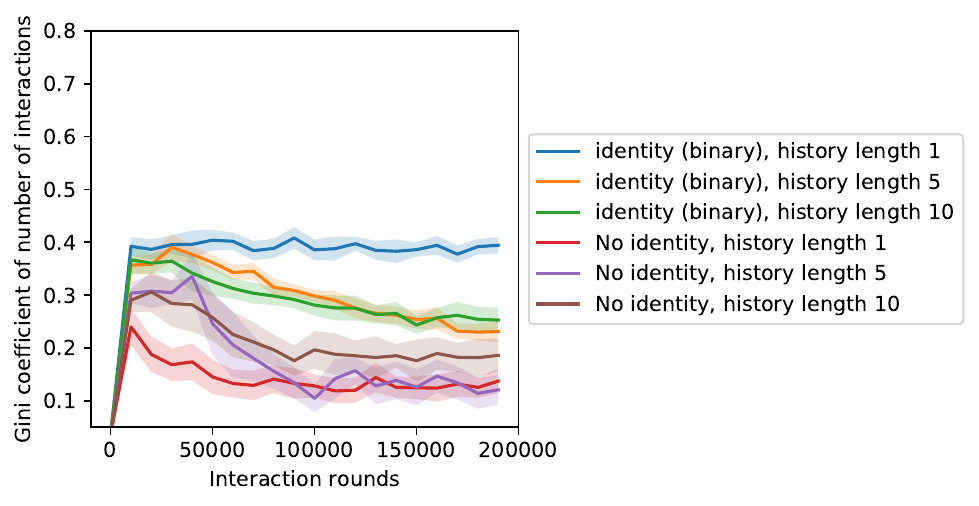}
    \caption{Dynamics of inequality (Gini coefficient) of partner selection between agents per 100 interaction rounds. Shaded areas show 95\% confidence interval across 20 runs.}
    \label{fig:gini-num}
\end{figure}

\begin{figure}[!htbp]
    \centering
    \includegraphics[width=0.4\textwidth]{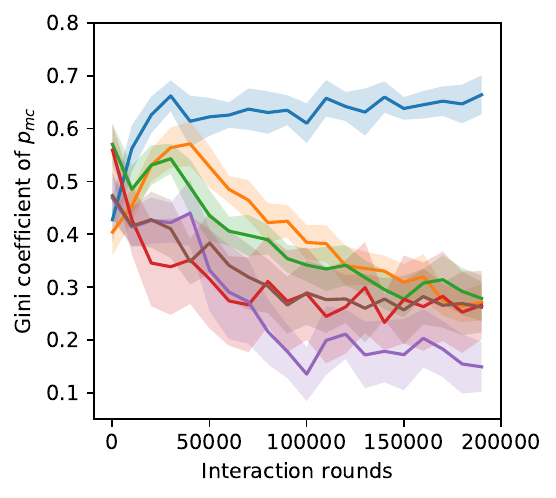}
    \includegraphics[width=0.4\textwidth]{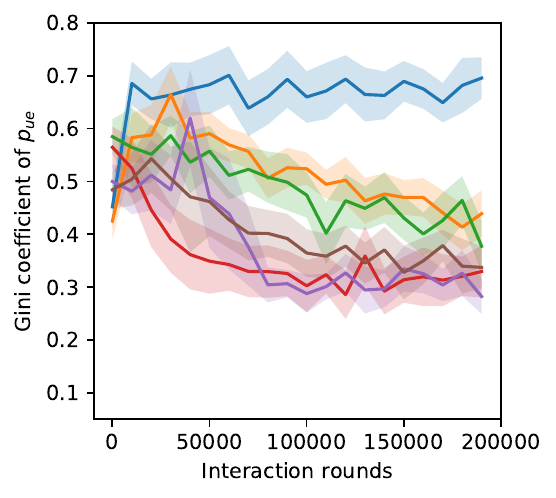}
    \includegraphics[width=0.4\textwidth]{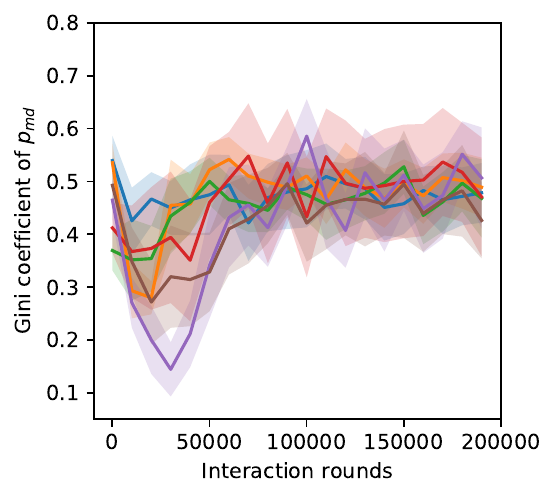}
    \includegraphics[
        width=0.4\textwidth,
        trim={0 -2cm 0 0}
    ]{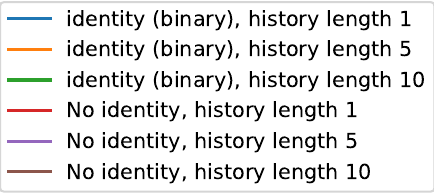}
    \caption{Dynamics of inequality (Gini coefficient) of proportion of different types of interactions. $p_{mc}$ (upper left), $p_{ue}$ (upper right) and $p_{md}$ (bottom left) denote the proportion of mutual cooperation, unilateral exploitation (\textit{sucker} defined in Section~\ref{sec:game}) and mutual defection in 100 interaction rounds respectively. Shaded areas show 95\% confidence interval across 20 runs.}
    \label{fig:gini-prop}
\end{figure}

We first discuss the results of experiments without opponent identity provided. In these cases, partner selection module calculates Q-value estimates purely from neighbours' action histories. When $l = 1$, action history only shows the action of the previous interaction round. Assuming that the action was not chosen by exploration, an action history with action $C$ implies that the assigned opponent's strategy is not defector. This means that agents with defective strategies cannot distinguish cooperators from retaliators, not guaranteeing their benefits from unilaterally exploiting the opponent. It results in a decrease of defective actions, while the proportion of mutual cooperation surges. Mutual cooperation becomes prevalent in cases with $l = 5, 10$ as well, but more slowly. One of the main reasons of delayed convergence is that longer action history enabled partner selection modules to target cooperators while avoiding selecting retaliators. Because retaliators change their actions based on the action history of their opponents, their own history is more likely to contain defection. The action history of cooperators mostly consists of cooperation, unless their strategies had changed recently or defection was selected by exploration. Because agents with defective strategies can select another cooperator as a victim when their previous target changes its behaviour, it takes more time for retaliating behaviour to suppress exploiting behaviour. The corresponding interaction patterns are visualised in Figure~\ref{fig:netplot-ps}. This is supported by Figure~\ref{fig:gini-num} and Figure~\ref{fig:gini-prop}, which show the inequality (Gini coefficient) dynamics between agents. The number of interactions and the proportion of different types of interactions in 100 interaction rounds are measured. Among experiments where identity information is not included, the inequality of the number of interactions, mutual cooperation and unilateral exploitation reaches its lowest level more slowly for $l = 5, 10$ than for $l = 1$. The steady increase of mutual cooperation is delayed until most of the agents with cooperative action history changes their behaviour to retaliator strategy, being ready to suppress the defective behaviour in the population.

\begin{figure}[!htbp]
    \centering
    \includegraphics[width=0.9\textwidth, trim={8cm 0.2cm 8cm 0}]{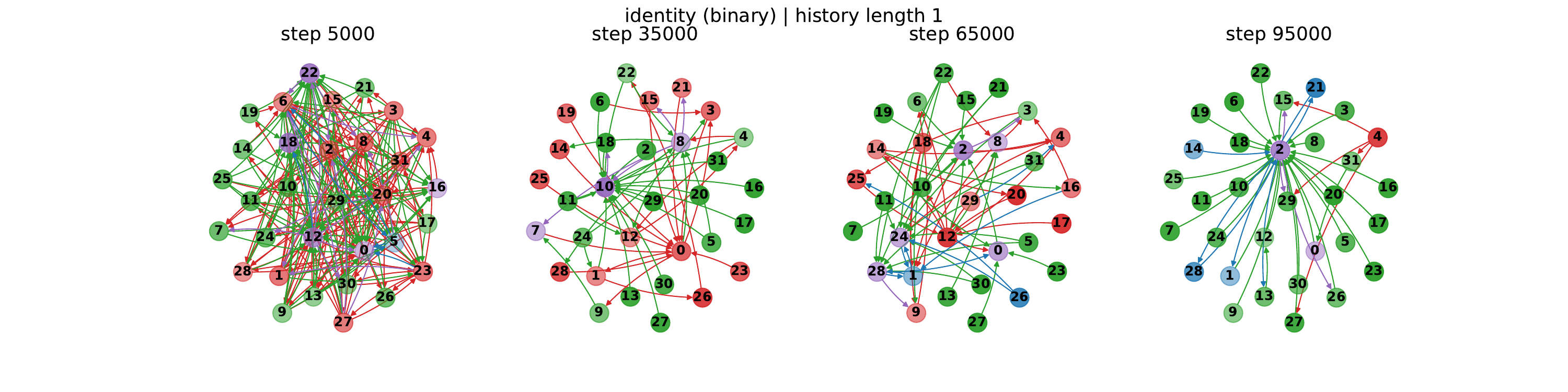}
    \includegraphics[width=0.9\textwidth, trim={8cm 0.2cm 8cm 0}]{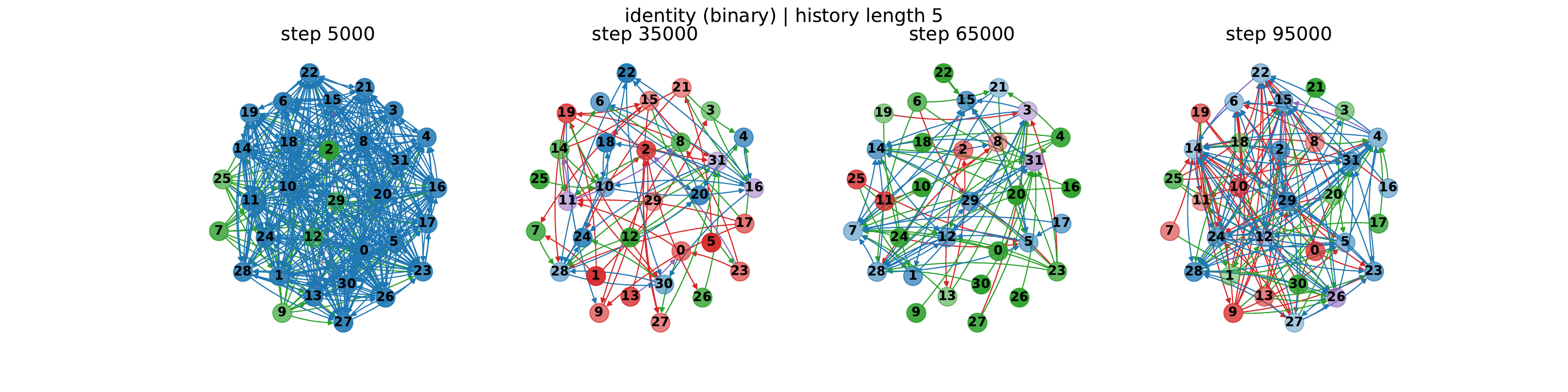}
    \includegraphics[width=0.9\textwidth, trim={8cm 0 8cm 0}]{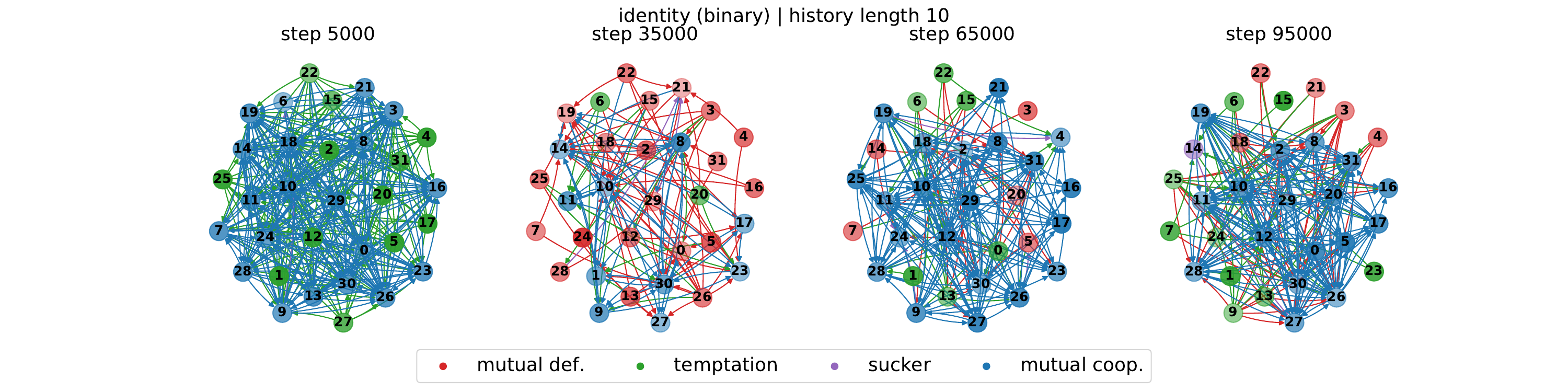}
    \caption{IPD interaction patterns with partner selection and binary-encoded identity ($l = 1, 5, 10$ from the top, respectively). Each graph shows results of one of the 20 runs in each experiment setting. When the length of the action history is 1, partner selections are highly concentrated. This concentration is alleviated as the history length gets longer, enabling more mutual cooperation to emerge in the population as experiment proceeds.}
    \label{fig:netplot-ps-id}
\end{figure}

In the experiments where opponent identity is included in state information, the overall dynamics differ most significantly from experiments without identity information with $l = 1$. Mutual cooperation is not prevalent even with the presence of partner selection mechanism. As both dilemma-playing network and partner selection module share the same format of state vector, dilemma-playing network chooses its action while also considering opponent identity. Each agent can learn to have different behaviours against opponents with different identity. This structure favours defectors over cooperators. Defective agents are not preferred during partner selection, making them play with only the agents they selected in an interaction round. Defectors are only required to learn to defect against their selection target. On the other hand, cooperative agents are more likely to be selected, encountering several others in an interaction round. Because they are likely to be exploited by defectors, cooperative agents have to learn to retaliate against all the opponents that are defecting. This process requires more learning steps than defective agents, because cooperative agents need to adapt their behaviour against multiple agents to avoid exploitation, while defective agents can benefit by learning to defect against a single agent. This unbalanced burden of learning is caused by introducing identity information and it hinders the emergence of cooperation. In cases with longer action histories $l = 5, 10$, additional information of opponents' behaviour make agents depend less on opponent identity. This lessens the concentration of partner selection while promoting the generalisation of agents' behaviour based on the opponents' action history, which enables the proliferation of mutual cooperation, as shown in Figure~\ref{fig:netplot-ps-id}. Figure~\ref{fig:gini-num} and Figure~\ref{fig:gini-prop} show that introducing identity information increased the inequality of the number of interactions, mutual cooperation and unilateral exploitation, while increasing action history length decreased the inequality. The variance of the proportion of interactions in Figure~\ref{fig:dyn-noid} and Figure~\ref{fig:dyn-id} is generally smaller in experiments with opponent identity than those without, because while the action history of agents continuously change throughout the experiment, identity information stays static and causes less variation in the choice of partner selection modules.

\section{Conclusion}
\label{sec:conclusion}

In this work, we have presented an analysis of the emergence of cooperation in a group of multiple agents interacting in the form of two-player IPD over a specified network structure. We have experimented with three different types of graphs, and have showed that in denser graphs and in those characterised by a shorter path length, mutual defection is more likely to spread. We have provided discount factor-based analysis to explain the relation between the results and the characteristics of graph structures. We also have analysed experiments with partner selection mechanism, altering the format of state vectors. We have considered scenarios composed of different lengths of action history, as well as binary-encoded identity information of the opponent. We have shown that longer action history slowed the emergence of mutual cooperation by enabling agents with defective behaviour to target cooperative agents while circumventing retaliators. We have experimentally demonstrated that providing agents with opponent identity results in lower level of mutual cooperation when insufficient information is provided in action history. Our work investigated the limited cases of IPD, which is one of the simplest games which does not affect the agents' location in the given graphs. We also only experimented with identity information that enables individual identification of the opponents. We leave the experiments with environments where graph topology can change temporally and, more in general, the study of sequential social dilemma \cite{leibo2017multiagentreinforcementlearningsequential,jaques2019socialinfluenceintrinsicmotivation,hughes2018inequityaversionimprovescooperation}, for future work. Investigation of learning dynamics of RL-based agents with different types of identity information, such as group identity, would be another promising topic for future research.

\FloatBarrier

\section*{Data and code availability}

The code is publicly available at \url{https://github.com/geronest/graph-ipd}.

\section*{Acknowledgements}

This work was supported by the UK Engineering and Physical Sciences Research
Council (EPSRC) under grant EP/X028569/1 (``Satisficing Trust in Human Robot
Teams'').

\bibliographystyle{unsrtnat}
\bibliography{references}

\end{document}